\documentclass[10pt,twocolumn,letterpaper]{article}
\usepackage[pagenumbers]{cvpr}

\definecolor{cvprblue}{rgb}{0.21,0.49,0.74}
\usepackage[pagebackref,breaklinks,colorlinks,allcolors=cvprblue]{hyperref}
\usepackage{graphicx}
\usepackage{booktabs}
\usepackage{multirow}
\usepackage{bm}
\usepackage[labelsep=period]{caption}
\title{Feed-Forward 3D Gaussian Splatting Compression with Long-Context Modeling}
\author{Zhening Liu$^{1}$\thanks{Equal contribution.}~~, Rui Song$^{1}$\footnotemark[1]~~, Yushi Huang$^{1}$, Yingdong Hu$^1$, Xinjie Zhang$^{1}$ \\
Jiawei Shao$^{1,2}$, Zehong Lin$^{1}$\thanks{Corresponding authors.}~~, Jun Zhang$^{1}$\footnotemark[2]\\
$^{1}$Hong Kong University of Science and Technology,\\
$^2$ Institute of Artificial Intelligence (TeleAI), China Telecom \\
{\tt\small \{zhening.liu,rrui.song,yhuangin,yhudj,xinjie.zhang\}@connect.ust.hk,} \\
{\tt\small shaojw2@chinatelecom.cn,}
{\tt\small\{eezhlin,eejzhang\}@ust.hk}
}

\begin{document}
\maketitle
\begin{abstract}
3D Gaussian Splatting (3DGS) has emerged as a revolutionary 3D representation. However, its substantial data size poses a major barrier to widespread adoption. 
While feed-forward 3DGS compression offers a practical alternative to costly per-scene per-train compressors, existing methods struggle to model long-range spatial dependencies, due to the limited receptive field of transform coding networks and the inadequate context capacity in entropy models.
In this work, we propose a novel feed-forward 3DGS compression framework that effectively models long-range correlations to enable highly compact and generalizable 3D representations.
Central to our approach is a large-scale context structure that comprises thousands of Gaussians based on Morton serialization. We then design a fine-grained space-channel auto-regressive entropy model to fully leverage this expansive context. Furthermore, we develop an attention-based transform coding model to extract informative latent priors by aggregating features from a wide range of neighboring Gaussians. Our method yields a $20\times$ compression ratio for 3DGS in a feed-forward inference and achieves state-of-the-art performance among generalizable codecs.
\end{abstract}    
\section{Introduction}
\label{sec:Intro}
Driven by the rapid development of 3D vision and spatial intelligence, the representation of 3D content has become an active research field in both academia and industry. Among various 3D representations, 3D Gaussian splatting (3DGS) \citep{kerbl20233d} stands out as a prominent technique due to its photorealistic visual quality and real-time rendering capability. It has found a wide range of applications, including simultaneous localization and mapping (SLAM) \citep{matsuki2024gaussianslam,cheng2025tlsslam}, virtual reality \citep{zheng2024gps,hu2024eva,liu2024dynamics}, and 3D content digitization \citep{he2025survey,sun2025virtual,bao20253dsurvey}. However, the superior representation capability of 3DGS relies on a large number of Gaussians, which poses significant challenges for transmission and storage. For example, representing a realistic scene can require storage on the order of several gigabytes (GB). This substantial overhead necessitates effective compression techniques for 3DGS. 

Most existing 3DGS compression methods \citep{fan2023lightgaussian,chen2024hac,wang2024contextgs,bagdasarian20253dgszip,ali2025compression} learn a compact representation from the original multi-view images through exhaustive per-scene per-train optimizations.
While these approaches achieve impressive compression ratios, the iterative optimization process often takes from minutes to hours, making them impractical for low-latency applications. Moreover, these methods cannot directly encode in-the-wild 3DGS data without referring to the original multi-view images, as the optimization depends heavily on these ground-truth inputs.
In contrast, generalizable compressors employ feed-forward neural networks to encode data.
This type of codec has been extensively studied in conventional data modalities, such as images \citep{balle2018variational}, videos \citep{lu2019dvc,li2021deep}, and point clouds \citep{gao2025deep}, but remains largely underexplored in 3DGS compression. To our knowledge, FCGS \citep{chen2024fast} is the only existing attempt at feed-forward 3DGS compression.
This method employs a multi-path entropy model and voxel grid-based context to build the codec, compressing 3DGS attributes through the inference of a trained neural network. It has demonstrated the feasibility of feed-forward 3DGS compression.

\begin{figure*}[t]
    \centering
    \includegraphics[width=0.93\linewidth]{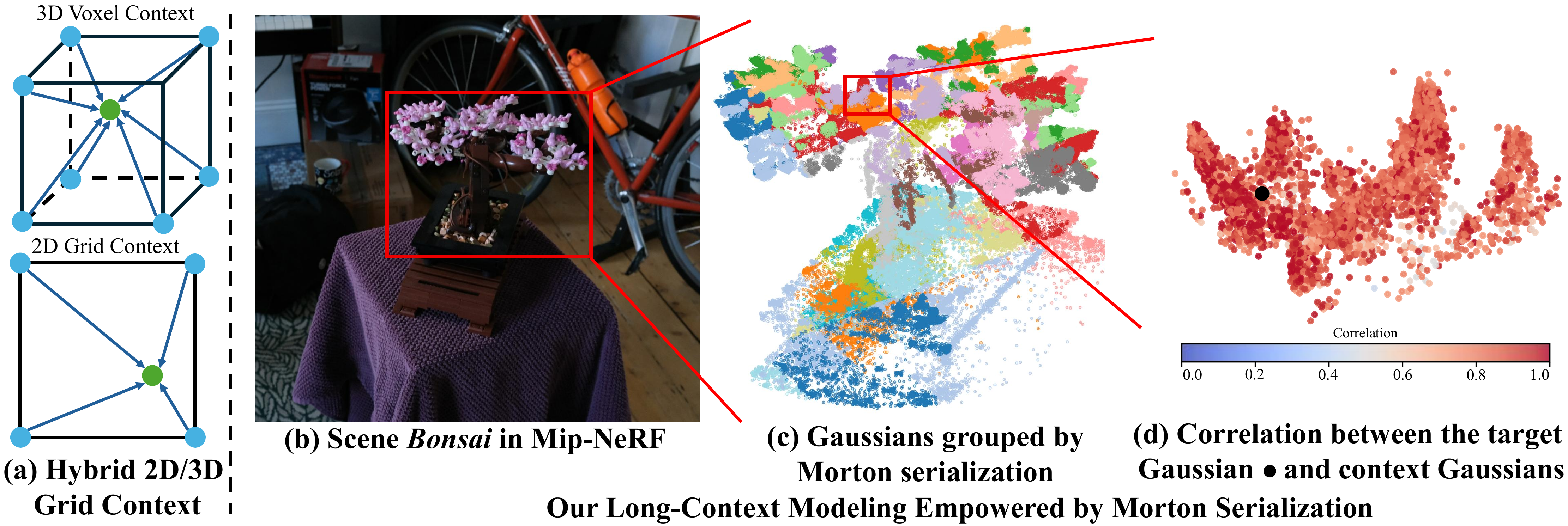}
    \caption{(a) Context illustration of the existing method \citep{chen2024fast}, which utilizes a limited receptive field that includes only several Gaussians. (b) Visualization of the \textit{Bonsai} scene. (c) Visualization of the corresponding 3DGS point cloud, where points are partitioned into Morton-serialized context windows. Points from the same context window are painted with the same color. 
    (d) Correlations of Gaussian attributes between a target point and other points within the same context window. Strong correlations persist between distant Gaussians. These visualizations illustrate the strong yet underexplored correlations among a large amount of 3DGS primitives, verifying the necessity of a large perceptive field and motivating our long-context modeling design.
    }
    \label{fig:context-visualization}
\end{figure*}

However, the design choices of FCGS remain suboptimal regarding the transform coding and context modeling. Specifically, it employs a multi-layer perceptron (MLP) to extract latent features from the original 3DGS representation, with its entropy model conditioned on a context that combines several neighbors in the voxelized 3D space and 2D projections. The MLP-based transform coding cannot effectively eliminate spatial redundancy since it processes each Gaussian independently and overlooks correlations among different Gaussians. The expressiveness of this latent representation is also limited due to the lack of spatial connectivity. In addition, the voxelized context entropy model only incorporates references within a narrow local vicinity, thus failing to capture the long-range dependencies in large-scale scenes. 
As shown in Fig. \ref{fig:context-visualization} (d), strong correlations persist across a substantial number of Gaussians. Thus, it is critical to enhance the long-range context modeling capabilities of both the transforms and the entropy models in order to boost the compression performance.

Nevertheless, discovering the long-range correlations among the unstructured 3DGS data remains a non-trivial problem. Since 3DGS representations are typically composed of up to millions of Gaussian primitives, it is impractical to compute dense pairwise attention across all primitives. Similarly, calculating correlations based on the $K$-nearest neighbors (KNN) algorithm requires exhaustive searches, which is computationally prohibitive as well.
Other solutions, such as computing correlations with randomly selected references, ignore the dependencies among Gaussian primitives and fail to provide informative contexts for compression.
To address these challenges, we propose \textbf{LocoMoco}, a novel method that achieves \textbf{lo}ng-\textbf{co}ntext \textbf{mo}deling for feed-forward 3DGS \textbf{co}mpression.
Our key idea is to scale the receptive field and context capacity to accommodate thousands of Gaussians guided by Morton serialization, and use attention mechanisms to effectively capture their long-range dependencies.
Our pipeline begins by traversing the 3DGS primitives into a 1D sequence via Morton serialization. This sequence is then partitioned into large-scale context windows, where Gaussians within the same window correspond to spatial neighbors, as depicted in Fig. \ref{fig:context-visualization} (c).
Subsequently, we devise self-attention modules that correlate Gaussians within each window to learn a compact and informative latent embedding. Then, we develop a large-scale context model that estimates the distributions of Gaussian attributes based on a fine-grained space-channel auto-regressive approach. To establish the spatial context, each window is first evenly split into two groups, after which the channel-wise context is incorporated to exploit correlations across different channels (i.e., Gaussian attributes). Finally, space, channel, and latent priors are integrated for a precise probability estimation.
To summarize, our contributions are as follows:
\begin{itemize}
    \item We propose a novel feed-forward 3DGS compression framework that effectively captures long-range dependencies in 3D scenes. To this end, we develop a large-scale context structure tailored for long-context modeling based on Morton serialization.
    \item We design a serialized attention architecture as the backbone of the transform coding network and context model. This design significantly expands the receptive field of the network, which effectively models the long-range dependencies in the extensive context. 
    \item We present a fine-grained space-channel context model for accurate distribution estimation. Guided by our serialization structure, it partitions the symbol sequence into closely correlated subgroups to fully exploit the dependencies among different Gaussians and attributes.
    \item Our method achieves state-of-the-art rate-distortion performance among feed-forward 3DGS compressors, surpassing baselines with a BD-Rate gain of around 10\%.
\end{itemize}

\section{Related Work}
\label{sec:Related}

\begin{figure*}[t]
    \centering
    \includegraphics[width=1.0\linewidth]{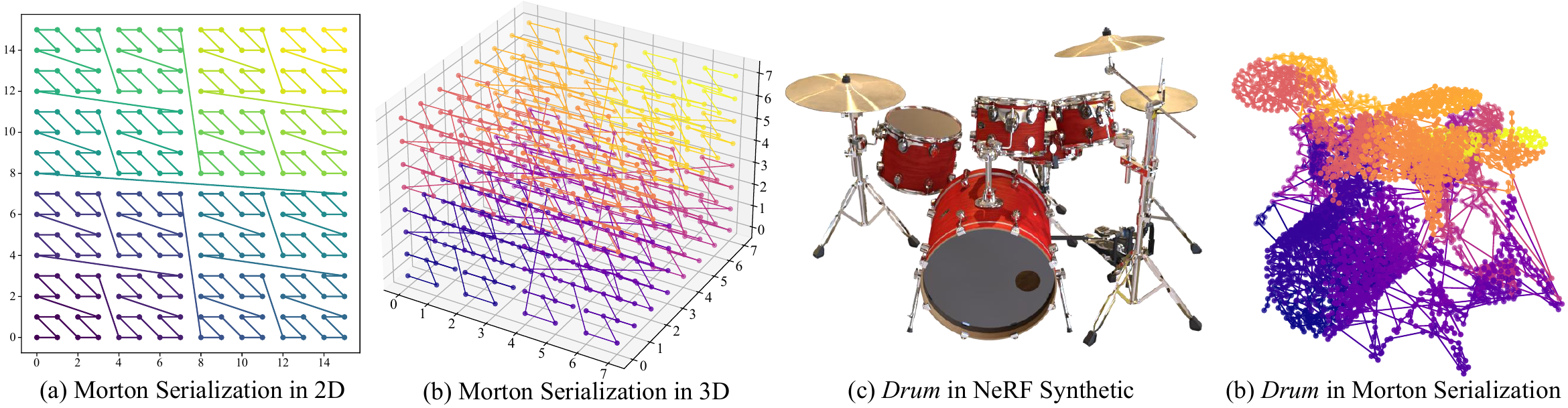}
    \caption{Visualizations of the Morton order in (a) 2D plane and (b) 3D space. We further exhibit the (c) visualization of the \textit{Drum} scene in the NeRF-Synthetic dataset \citep{nerf} and (d) the corresponding 3DGS point cloud, where points are colored according to their indices in the Morton-serialized sequence. 
    These visualizations illustrate that Morton serialization maintains the spatial proximity, which makes our context structure design effective.}
    \label{fig:morton-visualization}
\end{figure*}

\subsection{Optimization-based 3DGS Compression}
The substantial data volume of 3DGS representation hinders its wider-spread applications and motivates research interests in 3DGS compression \citep{bagdasarian20253dgszip,ali2025compression}. Existing 3DGS compression methods primarily adhere to the per-scene per-train paradigm, where an exhaustive optimization is required to yield a compact 3DGS representation based on multi-view image inputs. These methods can be broadly categorized into value-based and structure-based approaches. Value-based methods prune the less contributing elements from the representation \citep{fan2023lightgaussian,lee2024compact,ali2024trimming,zhang2025mega} and reduce numerical redundancies using quantization techniques \citep{girish2024eagles,niedermayr2024compressed,navaneet2024compgs,ali2025elmgs}. Structure-based methods simplify 3DGS into self-organized 2D planes \citep{morgenstern2023compact}, scaffold grids \citep{lu2024scaffold}, or other more compact data structures \citep{chen2024hac,wang2024contextgs,yang2024spectrally,liu2024compgs,liu2024hemgs,zhan2025cat,yang2025hybridgs}. Despite notable improvements in compression ratio, the practicality of these methods is limited by the time-consuming per-scene optimization and the necessary reliance on ground-truth multi-view images.

\subsection{Feed-forward Data Compression}
Different from the aforementioned methods that require per-scene optimization, most neural image, video, and point cloud compression models work in a feed-forward way. 
After training on large-scale datasets, these methods directly compress the data without further optimization. The feed-forward image and video compressors typically comprise a transform coding network and an entropy model \citep{balle2018variational, li2021deep}. The analysis transform projects the original data to a compact latent embedding, while the synthesis transform reconstructs the data from the latent conversely. The entropy model predicts the distributions of the latent embedding symbols to reduce the bitrate consumption, based on a context of previously decoded symbols \citep{minnen2018joint,minnen2020channel,he2022elic, jiang2023mlic,liu2024bidirectional,zhang2025camsic}.

Learning-based point cloud geometry compression methods use various context models to reduce the bitrate consumption for lossless compression. The orderless point cloud is first transformed into more regular data structures, such as octree and voxel. Then, octree-based methods predict the occupancy symbols of the octree nodes based on a context of previously decoded ancestor and sibling nodes \citep{huang2020octsqueeze,fu2022octattention,song2023efficient,song2023large}. Similarly, voxel-based methods estimate the voxel occupancy status based on decoded neighboring and low-resolution voxel grids \citep{wang2025versatile-geometry}. For attribute compression, hand-crafted or learned transforms are first employed to eliminate the redundancy within the data \citep{de2016compression,zhang2023scalable}. Then, a learning-based context model is adopted to further reduce the bitstream length for coding the transform coefficients or latent embeddings \citep{fang20223dac, wang2025versatile-intensity}.

Although the point cloud shares certain similarities with 3DGS representation, point cloud attribute compression techniques cannot be directly applied to 3DGS data. Point clouds commonly exhibit structured spatial organizations, where the attribute correlations are highly centralized in the narrow local neighborhood. As a result, point cloud attribute compressors can perform well even using localized receptive fields \citep{wang2025versatile-intensity, zhang2023scalable}. On the other hand, 3DGS employs high-dimensional anisotropic Gaussian primitives whose attributes are more irregular due to optimization. The Gaussian attributes maintain both local and long-range correlations induced by the view-dependent rendering and large-scale lighting effects. These properties highlight the importance of seeking correlated references in a more informative long-range context for 3DGS compression.
\section{Preliminaries}
\label{sec:prelimi}

\begin{figure*}[ht]
    \centering
    \includegraphics[width=0.99\linewidth]{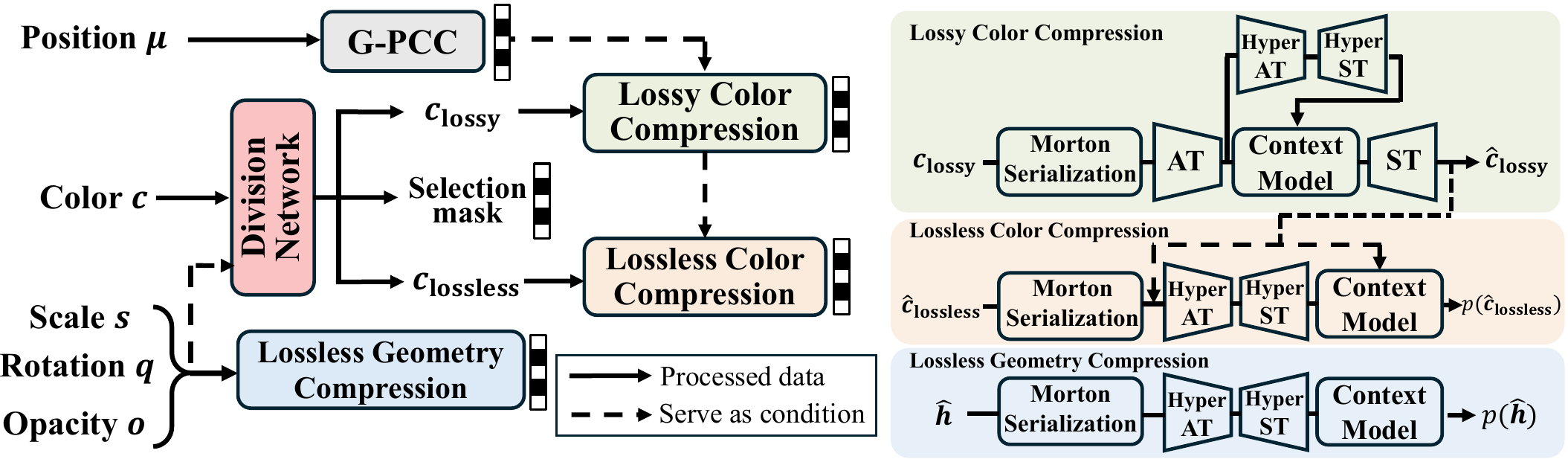}
    \caption{Overview of the proposed LocoMoco. The 3DGS attributes are compressed using different settings, catering for their individual properties. The architectures of each module are illustrated on the right hand side. Here, AT and ST denote the analysis and synthesis transforms, respectively. Both the transform coding network and the context entropy model have the long-context modeling capability.}
    \label{fig:pipeline-overview}
\end{figure*}

\subsection{3D Gaussian Splatting (3DGS)}
3DGS \citep{kerbl20233d} represents a 3D scene with a set of point-based primitives. Each Gaussian primitive is defined as:
\begin{equation} 
    \boldsymbol{g}(\bm{x}) = e^{-\frac{1}{2}(\bm{x}-\bm{\mu})^T \bm{\Sigma}^{-1} (\bm{x}-\bm{\mu})}, 
    \label{3dgs}
\end{equation}
where $\bm{\mu} \in \mathbb{R}^3$ and $\bm{\Sigma} \in \mathbb{R}^{3 \times 3}$ denote the position vector and the covariance matrix, respectively. The covariance matrix $\bm{\Sigma}$ can be decomposed into a rotation matrix $\mathbf{R} \in \mathbb{R}^{3\times3}$ and a scaling matrix $\mathbf{S} \in \mathbb{R}^{3\times3}$ as $\bm{\Sigma} = \mathbf{R}\mathbf{S}\mathbf{S}^T\mathbf{R}^T$,
where $\mathbf{R}$ and $\mathbf{S}$ are further represented by a quaternion $\bm{q} \in \mathbb{R}^{4}$ and a vector $\bm{s} \in \mathbb{R}^{3}$, respectively. 
Besides, each Gaussian contains an opacity property $o \in \mathbb{R}^1$ and a view-dependent color property (i.e., spherical harmonics) $\bm{c} \in \mathbb{R}^{48}$, which are incorporated into the rendering process as:
\begin{equation}C=\sum_{i=1}^N\bm{c}_i\alpha_i\prod_{j=1}^{i-1}(1-\alpha_j).
\end{equation}
Here, $\alpha_i$ denotes the transmittance of a Gaussian, which is calculated based on its opacity and covariance matrix. 
As listed above, a Gaussian primitive $\boldsymbol{g}$ is characterized by totally 59 parameters, and faithfully representing a 3D scene requires millions of Gaussian primitives. This leads to heavy storage burdens and motivates the need for effective and fast feed-forward 3DGS compression.

\subsection{Morton Code}
Morton code \citep{morton1966computer}, also known as Morton order or Z-order, is a commonly used technique in data serialization. It functions to sort the high-dimensional data according to the neighboring relationship in the high-dimensional space. 
Given a positive 3DGS coordinate $(x, y, z)$, which is quantized into $d$ bits, the Morton code $\pi$ is generated by interleaving the bits in the binary representation of $x$, $y$, and $z$:
\begin{equation}
\begin{aligned}
    \pi = \sum_{i=0}^{d-1} [& \text{Binary}(x, i) << (3i) + \text{Binary}(y, i) << (3i+1) \\
    & + \text{Binary}(z, i) << (3i+2)],
\end{aligned}
\end{equation}
where $\text{Binary}(u,i)$ extracts the $i$-th bit of $u$, and ``$<<$'' is the bit shifting operator. The Morton serialization rearranges the Gaussian primitives in the ascending order of Morton codes. Due to their similar Morton codes, neighboring Gaussians in the 3D space are placed nearby in the serialized sequence as well, as visualized in Fig. \ref{fig:morton-visualization}.
\section{Method}
\label{sec:method}
\subsection{Overview}
3DGS representation serves as an intermediate for novel view synthesis, and the rendering quality is sensitive to the perturbations on attribute values: minor deviations of Gaussian attributes can lead to substantial distortions in the rendered images. Motivated by existing methods \citep{chen2024fast}, we adopt a hybrid lossy-lossless compression backbone to yield high-fidelity rendering while achieving a compact data size. The geometric features (i.e., position $\boldsymbol{\mu}$, scale $\boldsymbol{s}$, rotation $\boldsymbol{r}$, and opacity $o$) are encoded losslessly, since the rendering quality is extremely sensitive to these components. Specifically, $\boldsymbol{\mu}$ is quantized into $\hat{\boldsymbol{\mu}}$ and compressed by an off-the-shelf point cloud codec G-PCC \citep{gpcc_codebase}. The other geometric features $\boldsymbol{h} = \left \{ \boldsymbol{s}, \boldsymbol{r}, o\right \}$ are discretized using learned quantization steps, and the quantized features $\hat{\boldsymbol{h}} = \left \{ \hat{\boldsymbol{s}}, \hat{\boldsymbol{r}}, \hat{o} \right \}$ are subsequently encoded via a neural context model. As for the color attributes $\boldsymbol{c}$, we first employ a lightweight division network to identify the perturbation-sensitive Gaussians, denoted by $\boldsymbol{c}_{\text{lossless}}$. These color attributes are then quantized into $\hat{\boldsymbol{c}}_\text{lossless}$ and compressed losslessly. Finally, the remaining color attributes, denoted by $\boldsymbol{c}_{\text{lossy}}$, are encoded via lossy compression to achieve a higher compression ratio.

The architecture of our LocoMoco is illustrated in Fig. \ref{fig:pipeline-overview}. It sequentially performs lossless geometry compression, lossless-lossy component division, lossy color compression, and lossless color compression. Each compression module first permutes the input 3DGS data via Morton serialization, which produces a 1D Gaussian primitive sequence $\boldsymbol{G} = \left \{\boldsymbol{g}_1, \cdots, \boldsymbol{g}_N \right \}$. Then, the derived sequence is partitioned into numerous context windows $\boldsymbol{x}$ of length $L$:
\begin{equation}
\boldsymbol{G} = \left \{ \boldsymbol{x}_1, \boldsymbol{x}_{2}, \cdots, \boldsymbol{x}_{k} \right \}, \quad \boldsymbol{x}_i = \left \{\boldsymbol{g}_{(i-1)L+1}, \cdots, \boldsymbol{g}_{iL} \right \}.
\end{equation}
Here, $k=\lceil N/L \rceil$ is the number of context windows, and the context length $L$ is set to 1024 in our implementation. Importantly, since $\boldsymbol{G}$ maintains the spatial neighboring relationships, our context partition guarantees that Gaussian primitives within the same window locate close to each other in the 3D space.
Afterwards, the lossless compressor extracts the hyper latent embedding $\boldsymbol{u}_i$ from $\boldsymbol{x}_{i}$ and predicts the probability distribution of $\boldsymbol{x}_i$ based on the quantized latent $\hat{\boldsymbol{u}}_i$ along with the space-channel context. The lossy compression model projects $\boldsymbol{x}_i$ to the latent feature $\boldsymbol{y}_i$, which is quantized into $\hat{\boldsymbol{y}}_i$ and compressed by another hyper-prior $\hat{\boldsymbol{z}_i}$ and the space-channel context. Finally, the synthesis transform network reconstructs Gaussian attributes from $\hat{\boldsymbol{y}}_i$. Both the transform coding network and the entropy model are built with attention blocks.

\begin{figure}[t]
    \centering
    \includegraphics[width=1.0\linewidth]{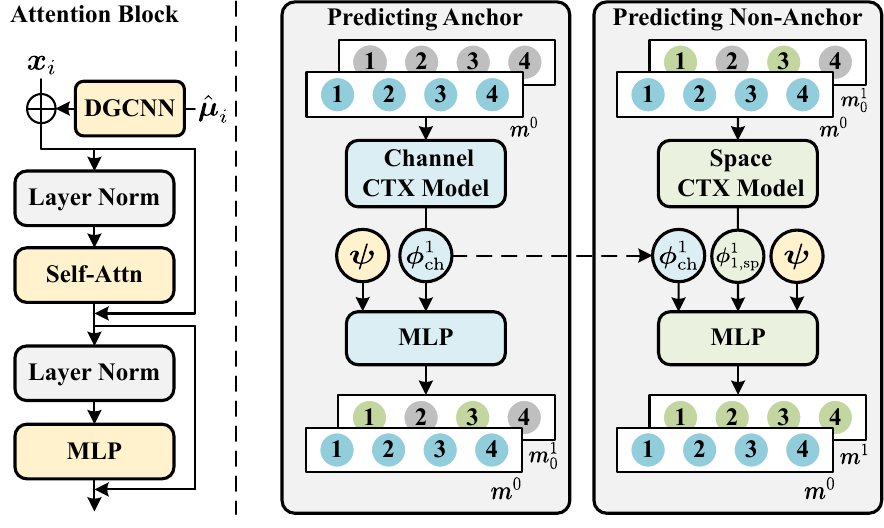}
    \caption{(Left) Architecture of the Morton serialized attention block. (Right) Workflow of the space-channel context model. We illustrate the coding of the subgroup $\boldsymbol{m}^{1}_0$ and $\boldsymbol{m}^{1}_1$ based on $\boldsymbol{m}^{0}$. Here, yellow nodes indicate the latent prior $\boldsymbol{\psi}$, blue nodes denote the channel context $\boldsymbol{m}^{0}$, and green nodes represent the decoded subgroups in $\boldsymbol{m}^{1}$.}
    \label{fig:context model}
\end{figure}

\subsection{Transform Coding} 
The transform coding model projects the original data to a compact latent space, which merges the duplicated patterns and removes the redundant information, and thus greatly reduces the data volume while supporting faithful reconstruction. Existing methods use an MLP-based transform coding network, where different Gaussian primitives are processed independently \citep{chen2024fast}. This method merely combines the attributes of each Gaussian primitive and fails to reduce the spatial redundancy by leveraging other Gaussians' features. Besides, the obtained latent features also tend to be less representative, because independent feature processing cannot analyze the spatial patterns.  

Inspired by 3D representation learning methods \citep{wu2024point}, we propose a transform coding module that utilizes a large-scale receptive field based on the attention mechanism. The structure of this attention block is shown in Fig. \ref{fig:context model} (Left). It first generates a positional encoding for each Gaussian using a dynamic-graph convolutional neural network (DGCNN) \citep{wang2019dynamic}, which extracts the geometric features of the local window $\boldsymbol{x}_i$ based on their positions $\hat{\boldsymbol{\mu}}_i$. Then, the self-attention layer computes dependencies among Gaussians within the same context window $\boldsymbol{x}_i$ and aggregates the features of different Gaussians based on the attention score. This self-attention layer facilitates the propagation of information among numerous Gaussian primitives, resulting in an informative latent embedding. This improvement is particularly important for lossy compression, as the reconstruction fidelity heavily depends on the expressiveness of the latent representation.

\subsection{Context Model} 
LocoMoco exploits a large-scale space-channel context to improve the probability density estimation.
Since the context model predicts different windows independently, without loss of generality, we define the entry of the context model as $\boldsymbol{m} = \left \{\boldsymbol{n}_{1}, \cdots, \boldsymbol{n}_{L} \right \}$. Here, $\boldsymbol{n}$ corresponds to Gaussian attributes or latent embeddings for lossless and lossy compression, respectively. The sequence $\boldsymbol{m}$ is partitioned into several slices along both spatial and channel dimensions, and different slices are predicted in serial to perform fine-grained space-channel context modeling. Specifically, $\boldsymbol{m}$ is first decomposed into two groups spatially:
\begin{equation}
    \setlength\abovedisplayskip{2pt}
    \setlength\belowdisplayskip{2pt}
    \begin{aligned}
    \boldsymbol{m}_{0} &= \left\{{\boldsymbol{n}}_{1}, {\boldsymbol{n}}_{3}, \cdots,  {\boldsymbol{n}}_{L -1} \right\}, \\
    \boldsymbol{m}_{1} &= \left\{{\boldsymbol{n}}_{2}, {\boldsymbol{n}}_{4}, \cdots,  {\boldsymbol{n}}_{L} \right\}.
    \end{aligned}
\end{equation}
This partition extracts elements with even and odd indices separately. As Morton serialization preserves the proximity in 3D space, the anchor $\boldsymbol{m}_{0}$ can be regarded as the spatial neighbors of the non-anchor $\boldsymbol{m}_{1}$. Thus, $\boldsymbol{m}_{0}$ provides a strong spatial context to predict $\boldsymbol{m}_{1}$. Then, each group is further partitioned into several subgroups along the channel dimension:
\begin{equation}
    \setlength\abovedisplayskip{2pt}
    \setlength\belowdisplayskip{2pt}
    \begin{aligned}
        \boldsymbol{m}^j_{0} &= \left\{{\boldsymbol{n}}^j_{1}, {\boldsymbol{n}}^j_{3}, \cdots,  {\boldsymbol{n}}^j_{L -1} \right\},\\
\boldsymbol{m}^j_{1} &= \left\{{\boldsymbol{n}}^j_{2}, {\boldsymbol{n}}^j_{4}, \cdots,  {\boldsymbol{n}}^j_{L} \right\},
    \end{aligned}
\end{equation}
where $\boldsymbol{n}_i^j$ indicates the $j$-th channel subgroup of $\boldsymbol{n}_i$.
The space-channel context model predicts symbol distributions by aggregating the space-channel context and latent features. The corresponding coding pipeline is illustrated in Fig. \ref{fig:context model}. Specifically, it first extracts a latent prior $\boldsymbol{\psi}$ from the latent $\hat{\boldsymbol{u}}$ (for lossless compression) or $\hat{\boldsymbol{z}}$ (for lossy compression). Then, it aggregates contextual features from the previously coded subgroups as:
\begin{equation}
    \setlength\abovedisplayskip{2pt}
    \setlength\belowdisplayskip{3pt}
    \begin{aligned}
        \boldsymbol{\phi}^{j}_{\text{ch}}= \sigma_{\text{ch}}&(\boldsymbol{m}^{<j}), \quad \boldsymbol{\phi}^j_{i, \text{sp}} = \sigma_{\text{sp}}(\boldsymbol{m}^{j}_{<i}), \\
        \boldsymbol{\phi}^j_{i} & = \sigma_{\text{ep}}(\boldsymbol{\phi}^j_{\text{ch}}, \boldsymbol{\phi}^j_{i, \text{sp}}).
\end{aligned}
\end{equation}
Here, $\boldsymbol{\phi}^j_{i}$ is the space-channel context for $\boldsymbol{m}^j_i$. The channel context model $\sigma_{\text{ch}}$ adopts self-attention blocks to extract features from previously coded channel subgroups $\boldsymbol{m}^{<j}$.
The space context model $\sigma_{\text{sp}}$ is only available for predicting the non-anchor $\boldsymbol{m}^{j}_1$. It uses self-attention blocks to introduce prior features from the decoded anchor $\boldsymbol{m}^j_0$. Here, features for anchor positions are derived by concatenating the latent features, channel context, and the embedding of $\boldsymbol{m}_0$, while the features for non-anchor positions are obtained by fusing the latent embeddings and channel context of $\boldsymbol{m}_1$. Finally, the latent prior $\boldsymbol{\psi}$ and space-channel context $\boldsymbol{\phi}^j_i$ are combined to predict the symbol distributions. Since the attention mechanism effectively models the long-range dependencies within the large-scale context $\boldsymbol{m}$, the available context capacity of the proposed model is significantly expanded compared to the voxel-based context \citep{chen2024fast}.

In addition, the decoded lossy color attributes $\hat{\boldsymbol{c}}_\text{lossy}$ serve as a context to predict the lossless ones $\hat{\boldsymbol{c}}_\text{lossless}$. Therefore, we concatenate the features extracted from $\hat{\boldsymbol{c}}_\text{lossy}$ and the latent priors of $\hat{\boldsymbol{c}}_\text{lossless}$ in the context model. Then, we perform Morton serialization and partition the context windows based on the concatenated sequence. The space-channel auto-regression is performed within $\hat{\boldsymbol{c}}_\text{lossless}$, and the prediction of $\hat{\boldsymbol{c}}_\text{lossless}$ benefits from the decoded features of $\hat{\boldsymbol{c}}_\text{lossy}$ through the attention blocks.

The entropy model uses a discretized logistic mixture \citep{salimans2017pixelcnn} to encode $\hat{\boldsymbol{h}}$ and $\hat{\boldsymbol{c}}_\text{lossless}$, while $\hat{\boldsymbol{y}}$ is encoded based on a normal distribution. The non-parametric density model \citep{balle2018variational} is adopted to encode $\hat{\boldsymbol{u}}$ and $\hat{\boldsymbol{z}}$. 

\subsection{Training}
Our pipeline is optimized using the rate-distortion (RD) objective. The overall training objective is formulated as:
\begin{equation}
\mathcal{L}_{\textrm{RD}} = \mathbb{E}_{\boldsymbol{G}}[D(\delta(\boldsymbol{G}), \delta(\hat{\boldsymbol{G}})) + \lambda R], 
\label{eq:RD-loss}
\end{equation}
where
\begin{align}
R = & \underbrace{\sum_{i,j}-\log p(\hat{\boldsymbol{h}}^j_i | \boldsymbol{\phi}^j_i, \hat{\boldsymbol{u}}_\text{geo}) - \log p(\hat{\boldsymbol{u}}_\text{geo})}_{\text{Lossless Geometry Rate}} \notag \\
+ & \underbrace{\sum_{i,j}-\log p(\hat{\boldsymbol{y}}^j_i | \boldsymbol{\phi}^j_i, \hat{\boldsymbol{z}}) - \log p(\hat{\boldsymbol{z}})}_{\text{Lossy Color Rate}} \notag \\
+ & \underbrace{\sum_{i,j}-\log p(\hat{\boldsymbol{c}}^j_{i, \text{lossless}} | \boldsymbol{\phi}^j_i, \hat{\boldsymbol{u}}_\text{color}, \hat{\boldsymbol{c}}_{\text{lossy}}) - \log p(\hat{\boldsymbol{u}}_\text{color})}_{\text{Lossless Color Rate}}.
\label{eq:rate}
\end{align}
Here, $\boldsymbol{G}$ represents the 3DGS representation for a scene sampled from the dataset, $\hat{\boldsymbol{G}}$ represents the recovered 3DGS attributes by merging the results of lossy and lossless compression, and $\delta$ indicates the rendering pipeline. The rate term $R$ comprises the bitrate for lossless geometry, lossy color, and lossless color components. The distortion $D$ is defined as a combination of the MSE and the SSIM loss computed between the rendered images $\delta(\boldsymbol{G})$ and $\delta(\hat{\boldsymbol{G}})$, which is averaged on six randomly chosen views during training. Besides, $\lambda$ is a hyper-parameter that controls the rate-distortion trade-off.
\section{Experiments}
\label{sec:exp}

\begin{figure*}[t]
    \centering
    \includegraphics[width=1.0\linewidth]{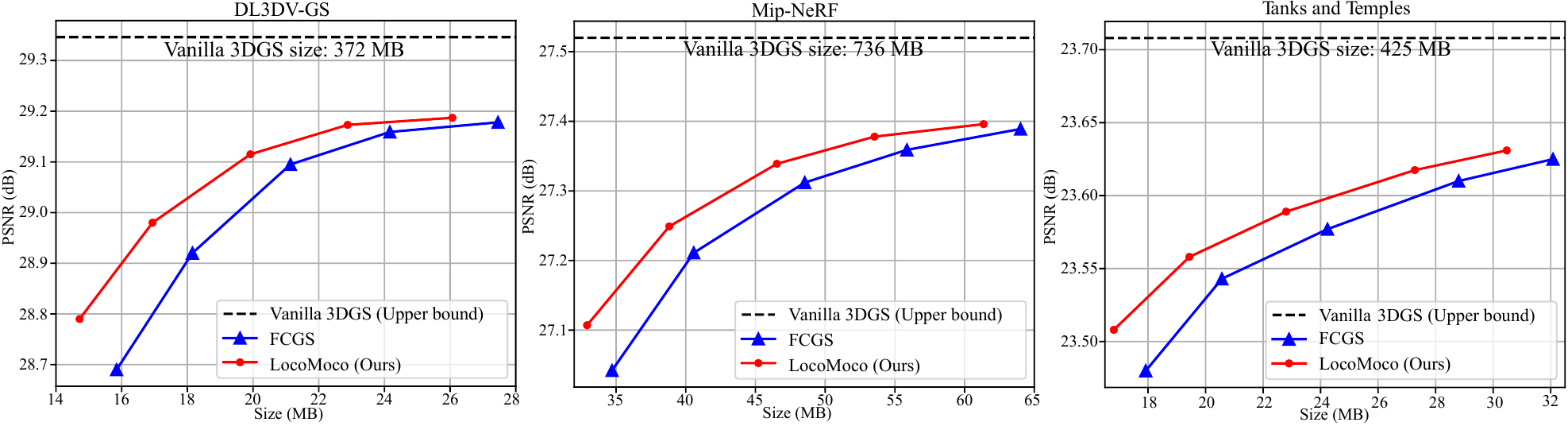}
    \caption{Rate-distortion curves evaluated on DL3DV-GS \cite{ling2024dl3dv}, Mip-NeRF \cite{barron2022mip} and Tanks \& Temples \cite{knapitsch2017tanks} datasets. The proposed LocoMoco yields significant compression compared to the vanilla 3DGS and consistently outperforms the baseline method.}
    \label{fig:RD-curve}
\end{figure*}

\subsection{Experimental Setup}
\noindent \textbf{Datasets.} We train the proposed LocoMoco on the DL3DV-GS dataset \cite{ling2024dl3dv,chen2024fast}, which contains 6,770 samples. Each sample comprises a set of multi-view images, corresponding camera poses, and 3DGS attributes. We follow the dataset division of FCGS \cite{chen2024fast}, where 100 scenes are preserved for evaluation while the others are used for training. Apart from these 100 samples, we conduct a comprehensive evaluation on the Mip-NeRF 360 \cite{barron2022mip} and Tanks \& Temples \cite{knapitsch2017tanks} datasets. We follow the vanilla 3DGS training pipeline \cite{kerbl20233d} to obtain the 3DGS representations from these datasets.

\begin{table*}[t]
  \centering
  \caption{(Left) BD-Rate gains relative to FCGS. (Middle) Encoding and decoding time evaluated on DL3DV-GS. (Right) Inference GPU memory consumption tested on DL3DV-GS. Both the peak and average GPU usage are reported.}
  \resizebox{\linewidth}{!}{%
    \begin{tabular}{cccccccc}
    \toprule
    \multirow{2}[4]{*}{Method} & \multicolumn{3}{c}{BD-Rate ($\downarrow$)} & \multicolumn{2}{c}{Time Consumption (s)} & \multicolumn{2}{c}{Inference GPU Memory (GB)} \\
    \cmidrule{2-4} \cmidrule{5-6} \cmidrule{7-8}
    & DL3DV-GS & Mip-NeRF & Tanks \& Temples & Encoding & Decoding & Peak Usage & Average Usage \\
    \midrule
    FCGS  & 0 & 0 & 0 & 15.65 & 11.33 & 41.8 & 29.1 \\
    LocoMoco (Ours) & -10.11\% & -9.39\% & -10.35\% & 11.33 & 12.89 & 44.9 & 34.5 \\
    \bottomrule
    \end{tabular}%
  }%
  \label{tab:BDBR-Time}%
\end{table*}%

\noindent \textbf{Implementation Details.} Our method is implemented using the Pytorch framework \cite{paszke2019pytorch}, aided with CompressAI \cite{begaint2020compressai} and FlashAttention \cite{dao2022flashattention} repositories. The latent feature $\hat{\boldsymbol{u}}$ is composed of $8$ and $48$ channels for lossless geometry and color compression, respectively. For lossy compression, the latent embedding $\hat{\boldsymbol{y}}$ includes $128$ channels.
Following FCGS, the batch size is set to $1$ for training. The parameter $\lambda$ is selected from $[1e-4,2e-4,4e-4,8e-4,16e-4]$ to switch among different rate-distortion trade-off levels. More details about the model architecture and implementation are presented in Appendix \ref{appendix-implementation}.

\noindent \textbf{Baselines.} To the best of our knowledge, FCGS \cite{chen2024fast} is the only existing exploration of feed-forward 3DGS compression. Therefore, LocoMoco is primarily compared against FCGS. We evaluate FCGS based on its official codebase and model weights. Besides, we conduct comparisons with various per-scene per-train 3DGS compression methods in Appendix \ref{appendix-per-scene-per-train}.

\noindent \textbf{Metrics.} The rate-distortion performance is measured using the storage size of the compressed 3DGS representation (in MB) and peak signal-to-noise (PSNR, in dB). We use the widely employed Bjøntegaard Delta Rate (BD-Rate) \cite{bjontegaard2001calculation} to compare the compression efficiency of different methods. The BD-Rate measures the relative data size savings between different methods under the same reconstruction quality, and thus a lower BD-Rate indicates better performance. More comparisons of the structural similarity index measure (SSIM) and learned perceptual image patch similarity (LPIPS) are provided in Appendix \ref{appendix-fidelity-metrics}. 

\subsection{Results}
\noindent \textbf{Quantitative Results.} Fig. \ref{fig:RD-curve} illustrates the rate-distortion performance of FCGS and LocoMoco on the DL3DV-GS, Mip-NeRF, and Tanks \& Temples datasets. Compared to the vanilla 3DGS representation that occupies hundreds of MB, our method greatly reduces the storage size and bears minor visual distortions (less than 0.5 dB drop in PSNR). Compared with FCGS, LocoMoco achieves consistently better rate-distortion performance across a wide range of bitrates among various datasets. 
Moreover, Table \ref{tab:BDBR-Time} suggests that the proposed method obtains around 10\% BD-Rate savings over FCGS, which is a significant improvement. 

The improvements stem from several advantageous designs. Our serialized-attention-based transform coding produces a compact and informative latent representation, which simultaneously reduces storage and improves reconstruction fidelity. Moreover, the proposed fine-grained space-channel context model effectively captures the long-range dependencies within the large-scale scenes, leading to precise probability estimation and reduced bitstream length.

Furthermore, we present the coding time of different methods in Table \ref{tab:BDBR-Time}. As a feed-forward 3DGS compressor, LocoMoco delivers a coding latency at the second level. It performs substantially faster than per-scene per-train compression methods whose runtime ranges from minutes to hours. Compared to FCGS, the superior compression performance of LocoMoco comes without compromising the coding efficiency. Notably, the proposed odd-even separated space context model requires only two iterations for spatial auto-regressive prediction, which is faster than the multiple rounds of grid interpolation used in FCGS. On the other hand, LocoMoco is built on a lightweight Transformer to pursue fast processing speed, where each attention block consists of only 1 or 2 self-attention layers, and the network width (i.e., channel dimension) is also squeezed to control the computational volume. Therefore, the usage of attention mechanism does not significantly affect the inference speed. Moreover, Morton serialization incurs negligible computational overhead, since it is  calculated by parallel bitwise operations. These designs ensure the efficiency of LocoMoco. 

In addition, we present the peak and average GPU memory usage in Table \ref{tab:BDBR-Time}. Although LocoMoco relies on self-attention computations, the results demonstrate that our method maintains a memory footprint comparable to FCGS. This is due to the following reasons. Firstly, LocoMoco compresses different context windows independently and in parallel. Thus, the memory consumption can be flexibly controlled by modifying the number of windows that are predicted in parallel. It also means that LocoMoco is applicable to arbitrarily large 3DGS representations without raising memory issues, because it can slice the 3DGS representation into smaller windows and encode different window batches in serial. Besides, the network width of LocoMoco is smaller than that of FCGS (e.g., 128 v.s. 256 for lossy color compressor), which also contributes to the memory usage reduction. We believe that the memory usage can be further reduced using the model quantization techniques for resource-constrained applications \cite{xiao2023smoothquant,li2023rain,huang2024tfmq,lin2024awq,huang2025qvgen}.

\begin{figure*}[t]
    \centering
    \includegraphics[width=1.0\linewidth]{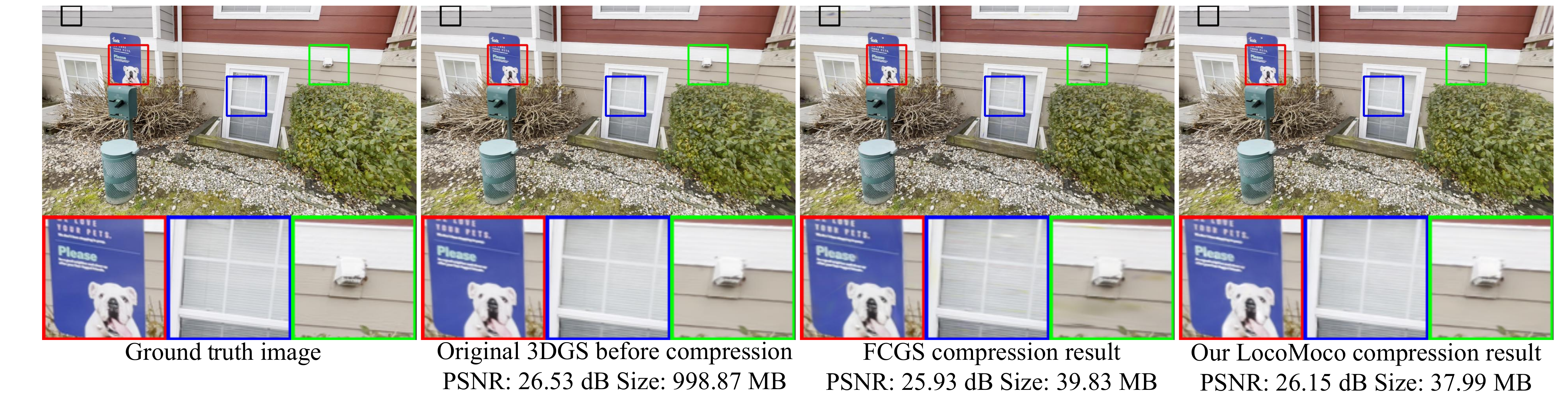}
    \caption{Qualitative results on DL3DV-GS, which compares the ground truth image and the rendering results of vanilla 3DGS, FCGS, and LocoMoco. The FCGS compressor causes blurred (see blue box) and distorted (see red, black, and green boxes) reconstruction. The proposed LocoMoco achieves high-fidelity reconstruction while reducing the storage size.}
    \label{fig:DL3DV-visualization}
\end{figure*}

\begin{figure}[t]
    \centering
        \includegraphics[width=0.85\linewidth]{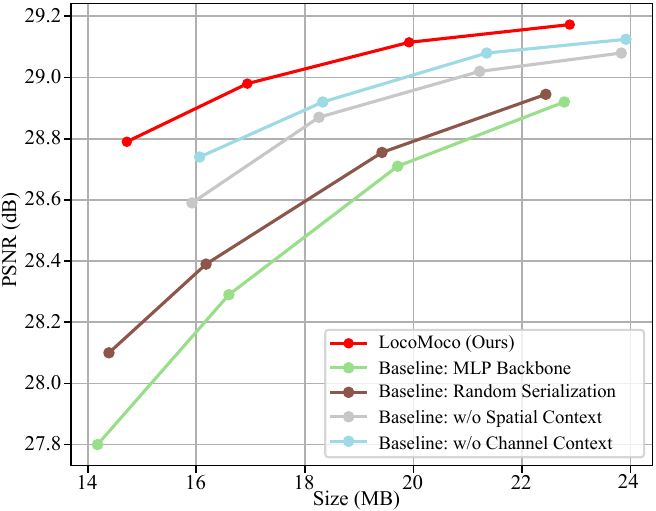}
        \caption{Rate-distortion curves of different ablation baselines evaluated on the DL3DV-GS.}
        \label{fig:ablation-main}
\end{figure}

\noindent \textbf{Qualitative Results.} In Fig. \ref{fig:DL3DV-visualization}, we visualize the ground truth image, the rendering result of vanilla 3DGS, and the rendered images from the compressed 3DGS representations produced by FCGS and LocoMoco. The proposed method yields a 26$\times$ data size reduction while preserving high-fidelity reconstruction. In contrast, the rendering result of FCGS exhibits blurred reconstructions and color variations. More visualizations are presented in Appendix \ref{appendix-visualizations}.

\subsection{Ablation Studies}
We conduct ablations experiments to analyze the contributions of the proposed serialized attention block and the spcae-channel context model.
More ablation studies are presented in Appendix \ref{appendix-ablation}.

\noindent \textbf{Morton Serialized Attention.} 
To verify the effectiveness of the serialized attention block, we conduct two variants here. The first baseline replaces all the attention blocks with MLPs. Notably, this model actually disables the space context because MLP cannot model the correlations among different Gaussians. The second baseline keeps the attention blocks while removing the Morton serialization operation. In this case, the attention is computed on a randomly permuted Gaussian sequence, which cannot guarantee that the Gaussian primitives are attended to their spatial neighbors. As shown in Fig. \ref{fig:ablation-main} and Table \ref{tab:ablation-main}, the MLP backbone baseline yields significantly worse rate-distortion performance. This experiment confirms the superiority of the attention network compared to MLP. Specifically, it demonstrates that the proposed attention model effectively captures the spatial dependencies, which are critical to enhance compression efficiency. The random serialization baseline also exhibits a large performance gap to the original model. It verifies that the attention model cannot perform well on weakly correlated sequences. In contrast, the attention model is effective on the Morton serialized sequence, where the spatial correlations are preserved.

\noindent \textbf{Space Context.}
To validate the importance of the space context, we implement a baseline where the spatial context is removed from lossless geometry, lossy color, and lossless color compression networks. Specifically, this model predicts the symbol distribution based on hyperpriors and channel contexts. As illustrated in Fig. \ref{fig:ablation-main} and Table \ref{tab:ablation-main}, applying the spatial context modeling leads to a BD-Rate gain of 18.45\%. This experiment verifies that the space context is quite informative, and the proposed space context model successfully utilizes these strong correlations.

\noindent \textbf{Channel Context}
Subsequently, we remove the channel context from the model while preserving the hyperprior and spatial references. Similarly, the channel context is removed from all networks. Results indicate that the channel context contributes to a significant BD-Rate improvement of 14.29\%, which demonstrates the effectiveness of the channel context model. In summary, both space and channel context is important to yield the satisfactory compression efficiency.

\begin{table}[t]
    \centering
        \caption{BD-Rate of the ablation baselines relative to the original LocoMoco model.}
        \begin{tabular}{cc}
            \toprule
            Method & \multicolumn{1}{l}{BD-Rate ($\downarrow$)} \\
            \midrule
            LocoMoco (Ours) & 0 \\
            Baseline MLP Backbone & +36.85\% \\
            Baseline Random Serialization & +30.62\% \\
            Baseline w/o Spatial Context & +18.45\% \\
            Baseline w/o Channel Context &  +14.29\% \\
            \bottomrule
        \end{tabular}
        \vspace{-5 pt}
        \label{tab:ablation-main}
\end{table}
\section{Conclusion}
In this work, we propose LocoMoco, a feed-forward 3DGS compression method that leverages informative long-range dependencies within 3D scenes. We propose a large-scale context structure tailored for 3DGS data using Morton serialization. Its context capacity reaches 1024 Gaussian primitives, which has been significantly expanded compared to existing voxel-based methods. Furthermore, we devise an attention architecture to effectively model the long-range dependencies. Moreover, we formulate a space-channel context model to fully exploit the contextual features conveyed by the large-scale context for precise probability density estimation. Extensive experiments demonstrate that LocoMoco achieves significant improvements in compression efficiency, while preserving the low-latency coding advantage of feed-forward 3DGS codecs.
{
    \small
    \bibliographystyle{ieeenat_fullname}
    \bibliography{main}
}
 
\newpage
\section*{Appendix}
\appendix
The appendix provides more implementation details in
Sec. \ref{appendix-implementation}, additional experimental results in Sec. \ref{appendix-more-experiments}, and extensive ablation studies in Sec. \ref{appendix-ablation}.

\section{Implementation Details}
\label{appendix-implementation}
\subsection{Model Architecture}
\begin{figure}[t]
    \centering
    \includegraphics[width=1.0\linewidth]{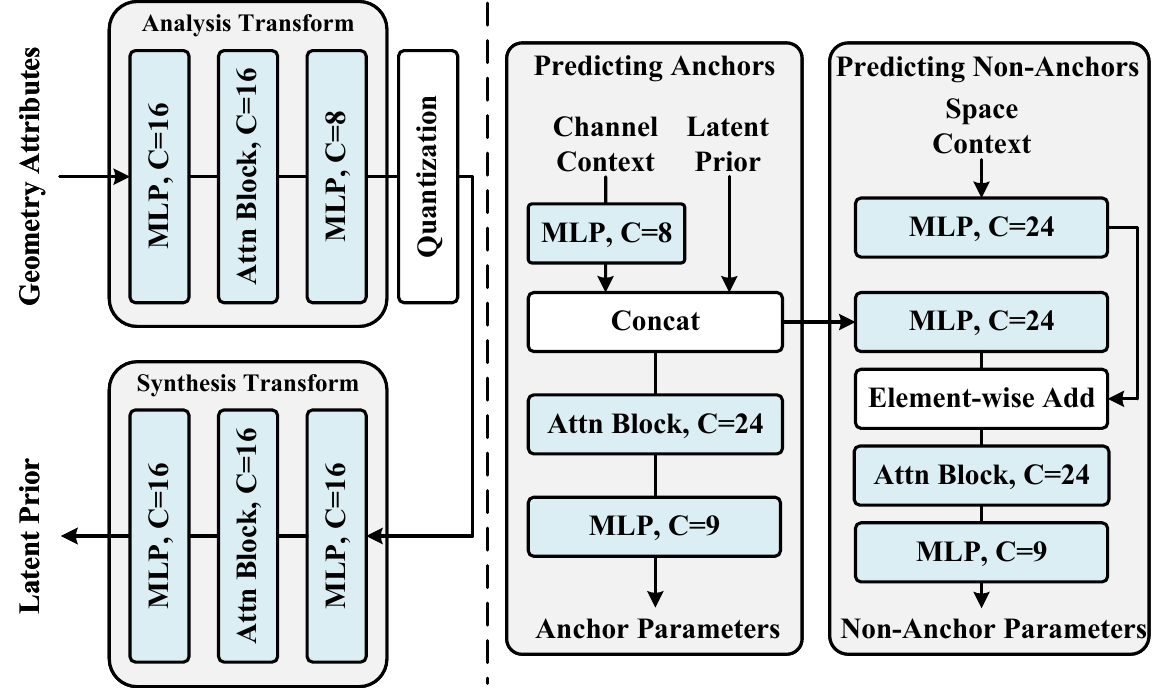}
    \caption{Structure of the lossless geometry compression model. The left and right parts illustrate the architectures of the transform coding network and the space-channel context model, respectively. Here, $C$ denotes the output dimension of each layer.}
    \label{fig:lossless-geo-model}
\end{figure}

\paragraph{Lossless Geometry Compression.} The lossless geometry compression model consists of hyper analysis transform, hyper synthesis transform, and space-channel context model. Its architecture is illustrated in Fig. \ref{fig:lossless-geo-model}. Both the analysis and synthesis transform coding networks are composed of $2$ MLPs and $1$ self-attention block. The latent vector $\boldsymbol{u}_\text{geo}$ includes $8$ channels. The space-channel context model consists of several MLPs and $2$ attention blocks. 

\begin{figure*}[t!]
    \centering
    \includegraphics[width=0.93\linewidth]{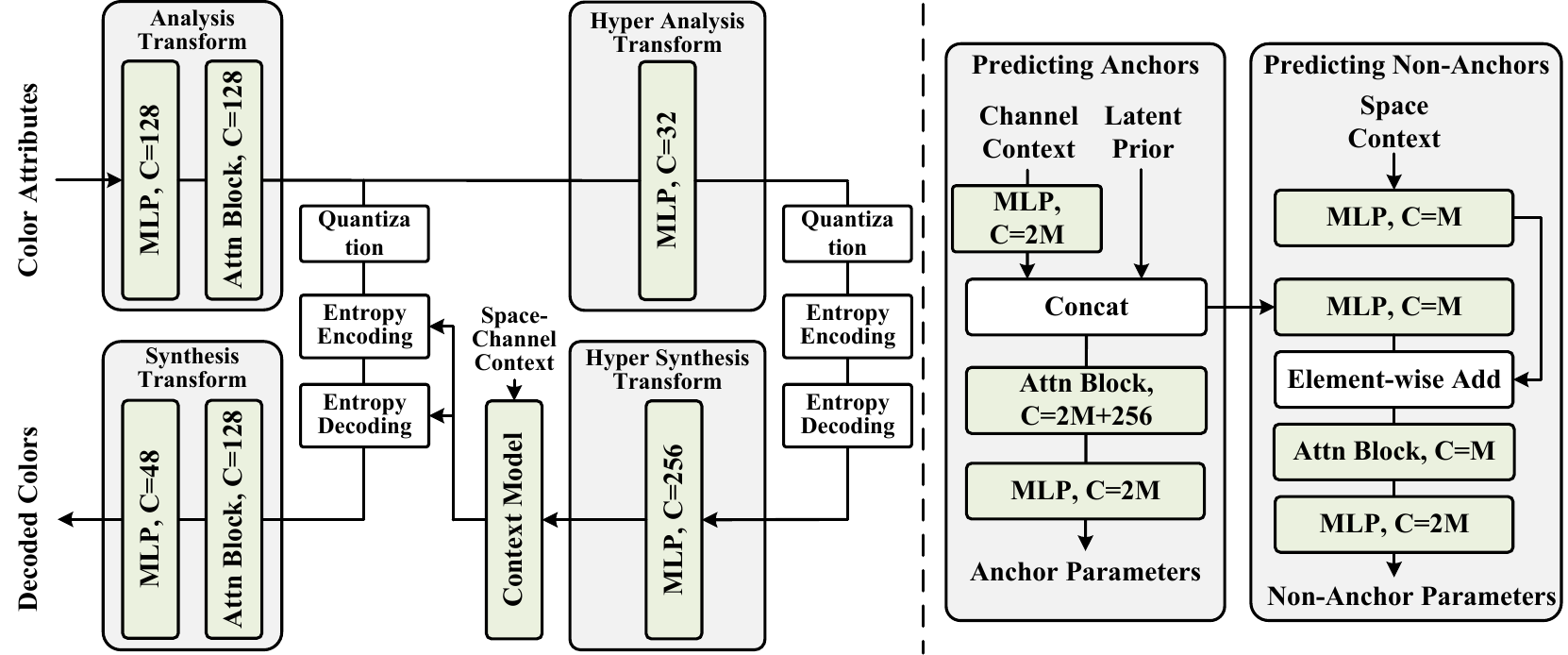}
    \caption{Structure of the lossy color compression model. The left and right part presents the architecture of the transform coding network and the space-channel context model, respectively.}
    \label{fig:lossy-color-model}
\end{figure*}

\paragraph{Lossy Color Compression.} The architecture of the lossy color compression network is shown in Fig. \ref{fig:lossy-color-model}. Its transform coding module and context model are basically consistent with the lossless compression pipeline. Since the lossy compressor needs to store more information in the latent embedding, the dimension of the latent features $\hat{\boldsymbol{y}}$ is set to $128$. The group size $M$ of the channel context model is $[32, 32, 64]$. The hyperprior embedding $\hat{\boldsymbol{z}}$ includes $32$ channels.

\begin{figure}[t]
    \centering
    \includegraphics[width=0.95\linewidth]{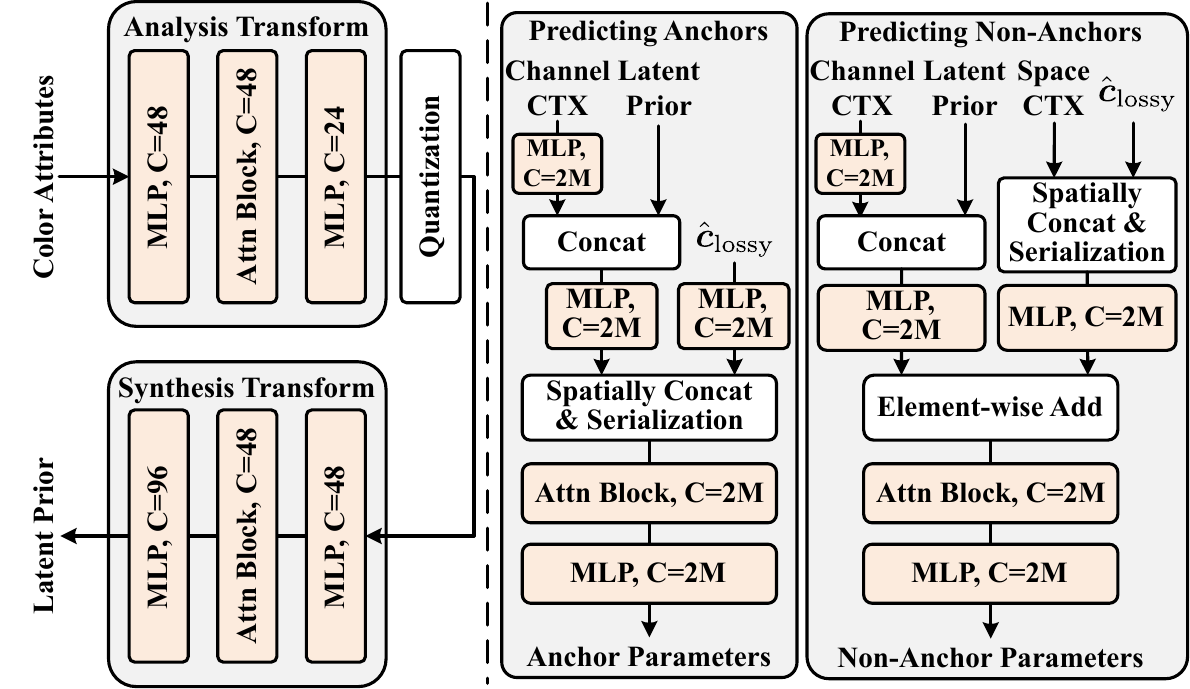}
    \caption{Structure of the lossless color compression model. The left and right parts separately illustrates the transform coding network and the context model.}
    \label{fig:lossless-color-model}
\end{figure}

\paragraph{Lossless Color Compression.} The structure of the lossless color compression module is shown in Fig. \ref{fig:lossless-color-model}. The latent embedding $\boldsymbol{u}_{\text{color}}$ has $48$ channels, and the channel group size is $[8, 8, 16, 16]$. In particular, we introduce the features of $\hat{\boldsymbol{c}}_\text{lossy}$ to predict $\hat{\boldsymbol{c}}_{\text{lossless}}$ by spatially connecting them in the same sequence. Then, we perform Morton serialization on the concatenated sequence and use an self-attention block to model the dependencies between $\hat{\boldsymbol{c}}_\text{lossy}$ and $\hat{\boldsymbol{c}}_{\text{lossless}}$.

\paragraph{DGCNN.} The attention block in LocoMoco employs a DGCNN \cite{wang2019dynamic} module to achieve positional encoding. The $k$-th DGCNN layer can be described as:
\begin{align}
\boldsymbol{f}^{k+1}_i &= \sigma_k([\boldsymbol{f}^{k}_i, \boldsymbol{f}^{k}_i - \boldsymbol{f}^{k}_j]), \quad \text{for } (i,j) \in \mathcal{E},
\end{align}
where $\sigma_k$ is an MLP, $\boldsymbol{f}^k_i$ is the feature of the $i$-th Gaussian, and $\mathcal{E}$ is a graph where each Gaussian is connected to its $K$ nearest neighbors in the feature space. After each layer, $\mathcal{E}$ is updated by recomputing the nearest neighbors in the new features. The first DGCNN layer receives the quantized Gaussian positions as input, i.e., $\boldsymbol{f}^0_i =\hat{\boldsymbol{\mu}}_i$. After several layers, DGCNN applies a maximum pooling layer to extract global features, and concatenates global and MLP features to produce the final embedding. In LocoMoco, we build the positional encoding module with $3$ DGCNN layers.

\paragraph{Division Network.} The division network, which is employed to split Gaussians for lossy and lossless compression, is built with one input embedding MLP, one attention block, and one output MLP. The channel dimension of the attention block is set to 128, and a binary selection mask is predicted based on the output of the attention block.

\subsection{Training Recipe}
In this subsection, we introduce the training pipeline of LocoMoco.
Directly balancing the rate-distortion trade-off by modifying the hyperparameter $\lambda$ is challenging, due to the complex interwoven effects among different compression submodules and the variable lossy-lossless division ratio. 
These instability issues (e.g., model collapse) have also been observed in prior work \cite{chen2024fast} when directly using $\lambda$ to control end-to-end training.  
Therefore, we adopt a multi-stage training strategy to stabilize the training process. Specifically, we first train each component individually, using a proxy loss function that does not involve $\lambda$. After the submodules are sufficiently initialized, we perform end-to-end tuning using the complete rate-distortion objective governed by $\lambda$.

As an initialization, the lossy color compression model and the division network are trained with a proxy loss function as follows:
\begin{align}
    \mathcal{L} = R + \lambda D(\delta(\boldsymbol{G}), \delta(\tilde{\boldsymbol{G}})) + w \cdot\frac{N_\text{lossless}}{N}.
\end{align}
With this training objective, we can vary the lossy-lossless division ratio by adjusting the hyperparameter $w$. To obtain five selections of $\lambda$, we train these networks with a target lossy compression ratio from $[85\%,80\%,75\%,70\%,65\%]$ for approximation. This strategy enables a separate training of the lossy color compressor without relying on $\lambda$. For lossless compression, we train the entropy model with the lossless geometry and color rate losses depicted in Eq. (\ref{eq:rate}), where the quantization steps are initialized from FCGS. Besides, we need to train the lossy and lossless color compression model sequentially, because the prediction of $\hat{\boldsymbol{c}}_{\text{lossless}}$ is based on the lossy compression results $\hat{\boldsymbol{c}}_\text{lossy}$. During the training of the lossless color model, the lossy color compressor and the division network are frozen. After the individual training, we finally fine-tune all modules in an end-to-end manner with the full loss in Eq. (\ref{eq:RD-loss}), where all network parameters are tuned to achieve a global rate-distortion optimization.

\begin{figure}[ht]
    \centering
    \includegraphics[width=1.0\linewidth]{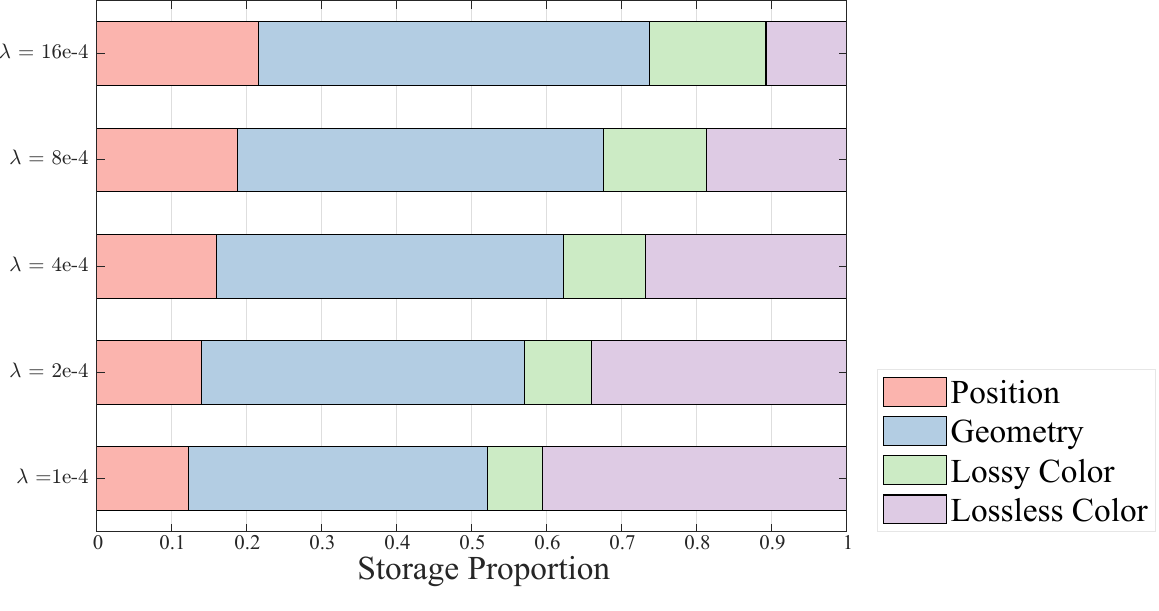}
    \caption{Proportion of each component's storage in LocoMoco, evaluated on the DL3DV-GS dataset. The division mask size, though included in the total file size, represents less than 1\% of the final result and is not displayed in this figure.}
    \label{fig:storage-allocation}
\end{figure}

\section{Supplementary Experimental Results}
\label{appendix-more-experiments}
\subsection{Fidelity Metrics}
\label{appendix-fidelity-metrics}
In the main paper, we provide the rate-distortion performance in terms of data size and PSNR. Here, we provide additional evaluation results for fidelity metrics, including SSIM and LPIPS. As shown in Table \ref{tab:appendix-visual-quality}, our method consistently achieves superior rate-distortion performance, maintaining fidelity similar to vanilla 3DGS while reducing storage overhead. Compared with FCGS, our method yields a higher visual quality with reduced storage overhead.

\begin{table*}[h]
  \centering
  \caption{Additional evaluation results in terms of PSNR, SSIM, LPIPS, and data size, evaluated on the DL3DV-GS dataset. The feed-forward compression methods achieve significant compression ratios on the vanilla 3DGS. Compared to FCGS, our method provides a higher visual quality with reduced storage overhead.}
    \begin{tabular}{ccccc}
    \toprule
    Method & PSNR (dB) $\uparrow$ & SSIM $\uparrow$ & LPIPS $\downarrow$ & Size (MB) $\downarrow$ \\
    \midrule
    Vanilla 3DGS & 29.344 & 0.8983 & 0.147 & 372.50 \\
    \midrule
    FCGS ($\lambda=0.0001$)  & 29.178 & 0.8966 & 0.149 & 27.96 \\
    FCGS ($\lambda=0.0016$)  & 28.692 & 0.8896 & 0.155 & 15.85 \\
    \midrule
    LocoMoco ($\lambda=0.0001$) & 29.186 & 0.8968 & 0.149 & 26.07 \\
    LocoMoco ($\lambda=0.0016$) & 28.789 & 0.8901 & 0.156 & 14.72 \\
    \bottomrule
    \end{tabular}%
  \label{tab:appendix-visual-quality}%
\end{table*}%

\begin{table*}[t!]
    \centering
    \caption{Time consumption of different 3DGS compressors. For feed-forward methods, both the encoding and decoding time are counted. The time consumptions of per-scene per-train methods consume minutes to hours for optimization and compression, whereas feed-forward generalizable codecs complete compression with a single inference pass, significantly reducing the processing time to 5\% of the former.}
    \begin{tabular}{cc|cc}
    \hline
    Method & Time for Compression (s) & Method & Time for Compression (s) \\
    \hline
    LightGaussian \cite{fan2023lightgaussian} & 978 & HAC \cite{chen2024hac} &  1800 \\
    Compact3D \cite{navaneet2024compgs}  & 1297 & ContextGS \cite{wang2024contextgs} &  5000 \\
    Compressed3D \cite{niedermayr2024compressed} & 873 & HEMGS \cite{liu2024hemgs} & 3600 \\
    SOG \cite{morgenstern2023compact} & 1068 & HAC++ \cite{chen2025hac++} & 2292 \\
    EAGLES \cite{girish2024eagles} & 518 & FCGS \cite{chen2024fast} & 26.98 \\
    Compact3DGS \cite{lee2024compact} & 938 & LocoMoco (Ours) & 24.22 \\
    \hline
    \end{tabular}
    \label{tab:appendix-compression-time}
\end{table*}

\subsection{Storage Proportion Allocation}
Fig. \ref{fig:storage-allocation} illustrates the storage proportion of each component in the LocoMoco, across various rate-distortion trade-offs. As the hyperparameter $\lambda$ decreases, the model tends to emphasize the low distortion aspect by increasing the storage for lossless color compression, which leads to better representation quality. 

\begin{figure*}[t!]
    \centering
    \includegraphics[width=0.99\linewidth]{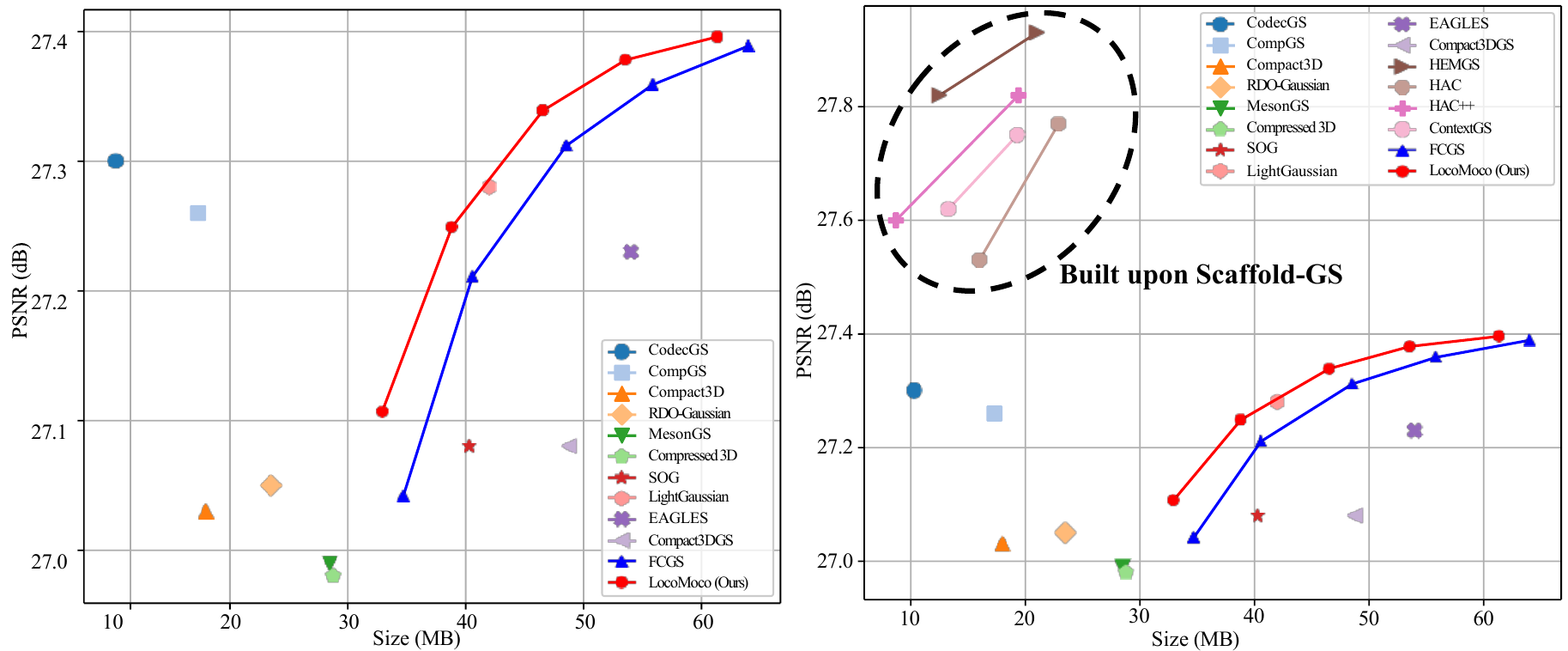}
    \caption{Rate-distortion performance comparison including the per-scene per-train compression baselines. The right figure includes the baselines built upon Scaffold-GS \cite{lu2024scaffold}, which achieve superior performance and distinguish themselves in the RD illustration.}
    \label{fig:all-compare-RD}
\end{figure*}

\subsection{Comparison with Per-Scene Per-Train Compression Methods}
\label{appendix-per-scene-per-train}
For the sake of completeness, we provide the rate-distortion performance comparisons against the per-scene per-train baselines in Fig. \ref{fig:all-compare-RD} and Table \ref{tab:appendix-compression-time}. We compare LocoMoco with a wide range of baselines, including HEMGS \cite{liu2024hemgs}, ContextGS \cite{wang2024contextgs}, HAC \cite{chen2024hac}, HAC++ \cite{chen2025hac++}, CodecGS \cite{lee2025compression}, CompGS \cite{liu2024compgs}, Compact3D \cite{navaneet2024compgs}, RDO-Gaussian \cite{wang2024end}, MesonGS \cite{xie2024mesongs}, Compressed3D \cite{niedermayr2024compressed}, SOG \cite{morgenstern2023compact}, LightGaussian \cite{fan2023lightgaussian}, EAGLES \cite{girish2024eagles}, Compact3DGS \cite{lee2024compact}, and FCGS \cite{chen2024fast}. 

Please note that among all these methods, only FCGS \cite{chen2024fast} and our LocoMoco are generalizable feed-forward codecs. As expected, the per-scene per-train methods achieve the better rate-distortion performance, since these methods overfit to each scene and exploit the ground truth multi-view images, while the feed-forward methods are trained towards the generalizable compression capability. However, as shown in Table \ref{tab:appendix-compression-time}, these pre-scene per-train methods require exhaustive optimization iterations to compress each 3DGS sample, which greatly slows down the coding speed. In contrast, feed-forward 3DGS codecs support much faster processing, because the computational overhead of network inference is much lower than that of training.

\subsection{Compressing 3DGS representations from Per-Scene Per-Train Methods}
\begin{figure}[t]
    \centering
    \includegraphics[width=1.0\linewidth]{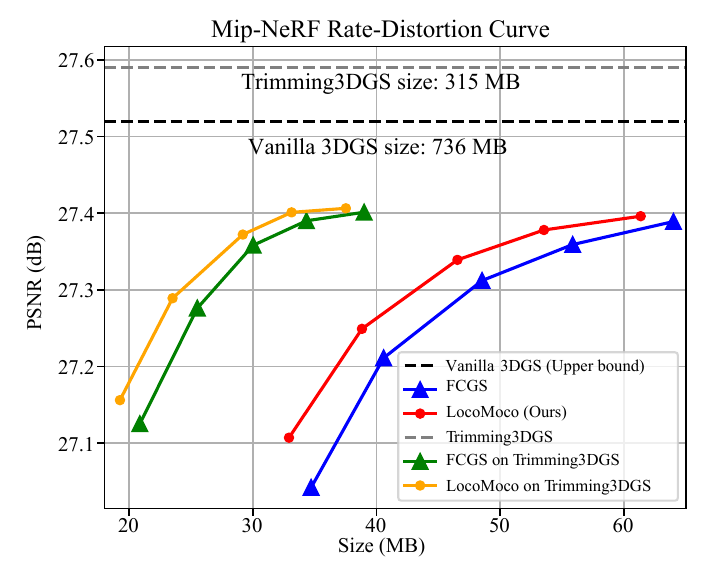}
    \caption{Rate-distortion performance comparison of applying the feed-forward codecs on per-scene per-train 3DGS representations, with the results evaluated on Mip-NeRF dataset.}
    \label{fig:compress-trimming}
\end{figure}

Our feed-forward 3DGS codec is applicable to any 3DGS data which have the canonical input format. Therefore, we explore applying this generalizable compression method to further compress the 3DGS representations obtained from the pruning-based per-scene per-train compression methods, e.g., Trimming3DGS \cite{ali2024trimming}. Trimming3DGS yields a compressed 3DGS representation by iteratively pruning low-contribution Gaussians and fine-tuning the remaining ones. We directly apply our feed-forward compressor to encode the 3D Gaussians acquired from this per-scene per-train approach without any fine-tuning. Results on the Mip-NeRF dataset are shown in Fig. \ref{fig:compress-trimming}.
Although the 3DGS samples produced by Trimming3DGS differ from the vanilla 3DGS data, our method still excels in compressing these out-of-distribution samples. This advantage probably stems from the long-context modeling capability of LocoMoco, which consistently captures the long-range dependencies in the pruned 3DGS representations and facilitates effective compression by removing the redundancy in the original data. 

\subsection{Effectiveness of Morton Serialization}
In the proposed LocoMoco framework, the long-context modeling is built by grouping Gaussian primitives into context windows following Morton serialization. In this subsection, we investigate whether the Morton serialization can preserve the spatial organization and establish strong correlations among Gaussian primitives. 

\begin{figure*}[h]
    \centering
    \includegraphics[width=0.98\linewidth]{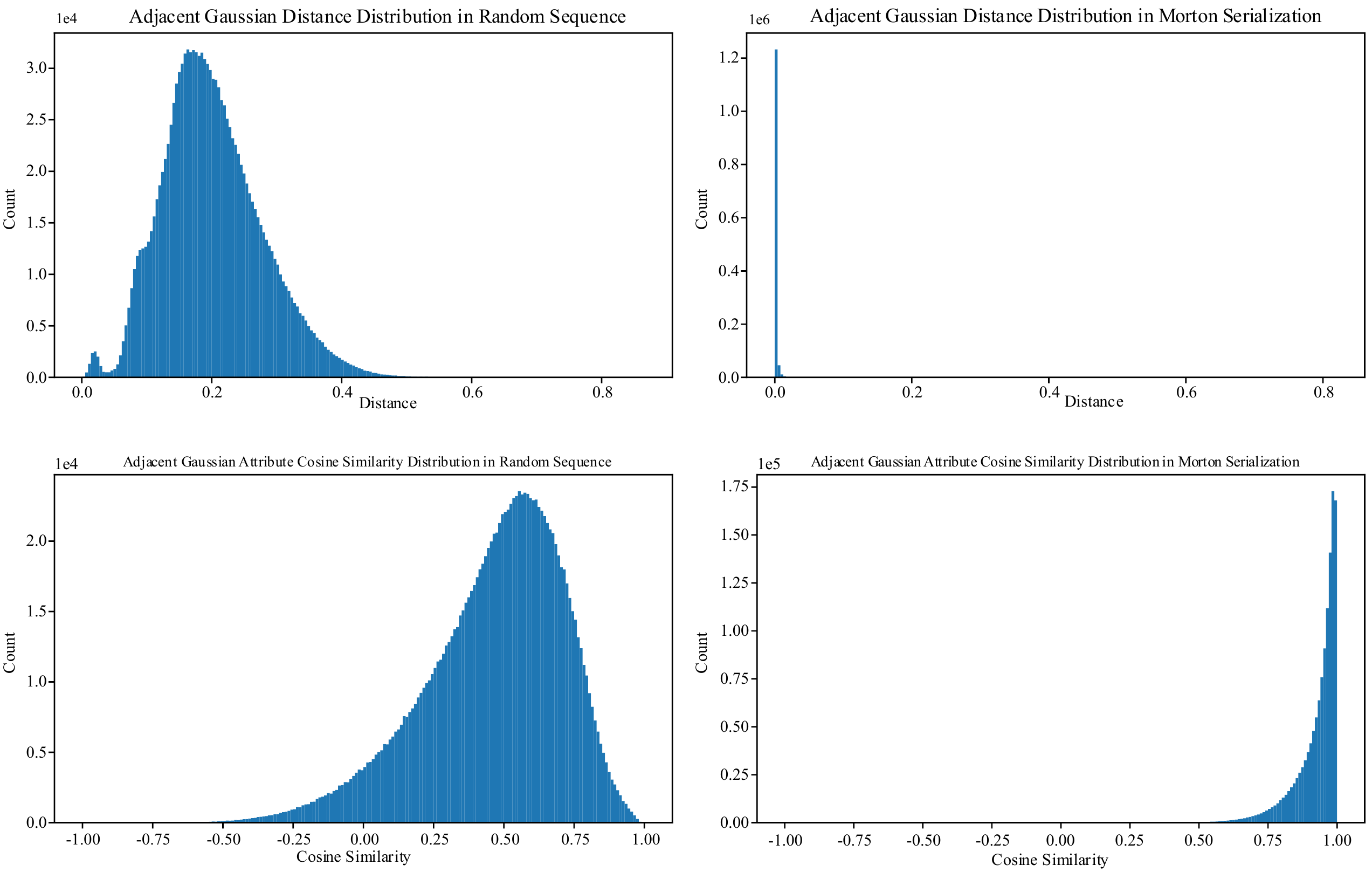}
    \caption{Histogram of the 3D distance (Top) and the cosine similarity (Bottom) between Gaussian primitives and their neighbors in the randomly permuted sequence and the Morton serialized sequence. The distances are normalized by the scene bounding box to $[0,1]$. The cosine similarities are calculated based on primitives' attributes and range from $[-1,1]$.
    In the Morton serialized sequence, neighboring Gaussian primitives exhibit strong spatial organization (Top right) and correlations (Bottom right). These results verify the effectiveness of our design choice of building long-context modeling based on Morton serialization.}
    \label{fig:appendix-morton-distribution}
\end{figure*}

Specifically, we compute the 3D distance and correlation between Gaussian primitives and their neighbors in the traversed sequence. 
In Fig. \ref{fig:appendix-morton-distribution} (Top), we compare the distributions of 3D distances between adjacent Gaussians in the randomly permuted sequence (Top left) and the Morton serialized sequence (Top right). We observe that the 3D distances between adjacent Gaussians under Morton serialization present a sharply concentrated distribution near 0, indicating that spatially neighboring Gaussians are placed close to each other under Morton serialization as well. In contrast, the randomly permuted sequence displays a relatively uniform distribution. This verifies that Morton serialization effectively maintains spatial relationships and enables our long-context modeling to capture the neighborhood contexts.
Moreover, in Fig. \ref{fig:appendix-morton-distribution} (Bottom), we visualize the distributions of cosine similarities between adjacent Gaussians in the random sequence (Bottom left) and Morton serialized sequence (Bottom right). The Morton serialization establishes significant correlation patterns, where most adjacent pairs show strong similarities close to 1, while the random permutation shows weak correlations. This comparison confirms that Morton serialization provides strong feature coherence for the context window attention computation, thereby contributing to the improved compression performance.

\subsection{Visualizations}
\label{appendix-visualizations}
\begin{figure*}[h]
    \centering
    \includegraphics[width=1.0\linewidth]{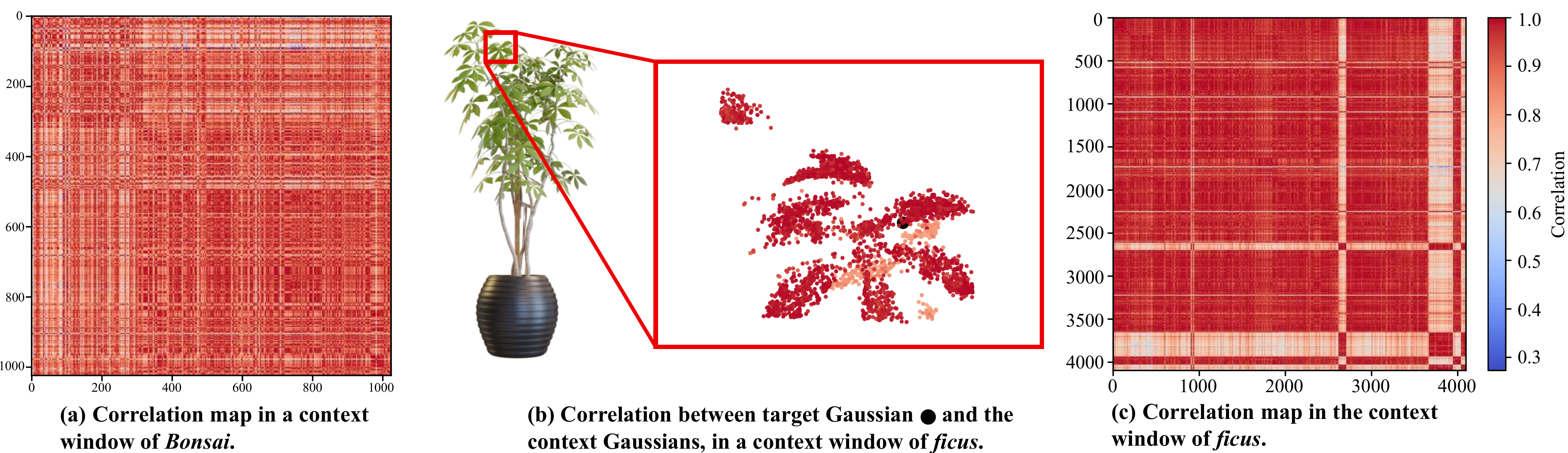}
    \caption{(a) Correlation map between the Gaussians in one context window of scene \textit{Bonsai}. (b) Correlation between one target Gaussian and the others in a same context window in scene \textit{ficus}. (c) Correlation  map between the Gaussians in one context window of scene \textit{ficus}.}
    \label{fig:ficus-visualization}
\end{figure*}

\begin{figure*}[h]
    \centering
    \includegraphics[width=1.0\linewidth]{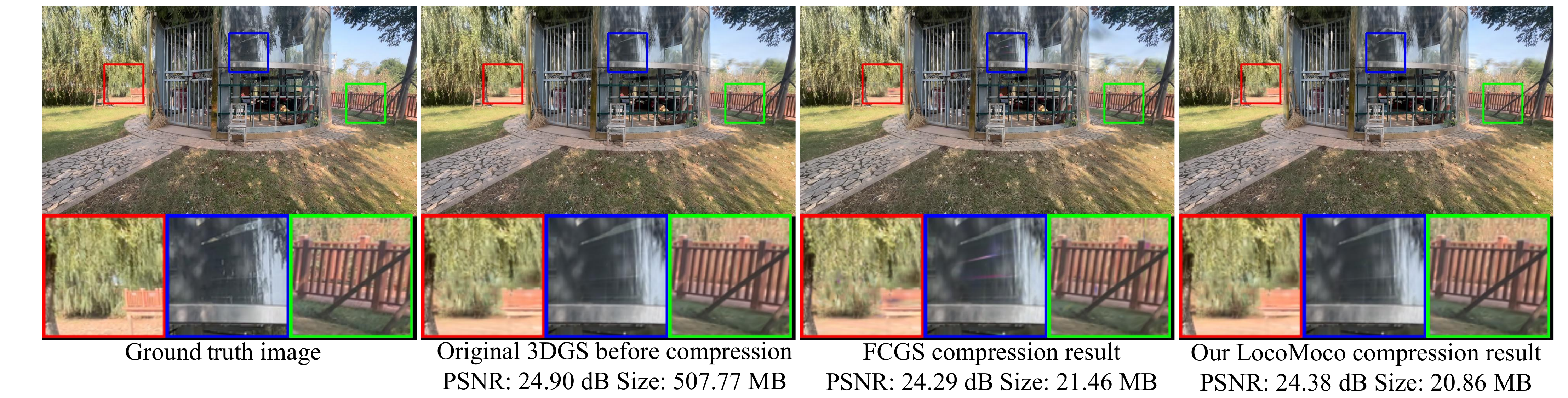}
    \caption{Qualitative results on DL3DV-GS, compared with the ground truth image and the rendering results of vanilla 3DGS, FCGS, and LocoMoco. Our proposed LocoMoco achieves high-fidelity reconstruction while reducing the storage size.}
    \label{fig:appendix-visualize}
\end{figure*}

As an extension of Fig. \ref{fig:context-visualization}, we include the correlation map between the Gaussians in a context window in Fig. \ref{fig:ficus-visualization} (a). We also illustrate another example of the scene \textit{ficus} in Fig. \ref{fig:ficus-visualization} (b)(c), where strong correlations can be observed in the region of leaves. These visualizations effectively verify the importance of long-context modeling in 3DGS compression. Moreover, we include a visualization comparison in Fig. \ref{fig:appendix-visualize}. The results show that our method delivers a more faithful reconstruction of the vanilla 3DGS representation compared to existing baselines, achieving color consistency and high-fidelity geometric details.

\begin{table*}[h!]
  \centering
  \caption{BD-Rate relative to the original LocoMoco model (where context length $L=1024$). The BD-Rate quantifies the additional storage required by baseline variants to achieve the same reconstruction quality as the original model.}
    \resizebox{\linewidth}{!}{
    \begin{tabular}{cccccccc}
    \toprule
    \multirow{2}[2]{*}{\textbf{Method}} & LocoMoco & \multicolumn{4}{c}{Baselines on context window size} & \multicolumn{2}{c}{Baselines on positional encoding} \\
    \cmidrule{3-8}
     & (Ours) & Context-2048 & Context-512 & Context-128 & Context-16 & Sinusoidal PE & DGCNN w/ feature \\
    \midrule
    \textbf{BD-Rate} ($\downarrow$) & 0 & -1.79\% & + 4.05\% & + 7.97\% & + 27.18\% & + 7.93\% & -0.52\% \\
    \bottomrule
    \end{tabular}%
    }%
  \label{tab:appendix-ablation}%
\end{table*}%

\section{Extensive Ablation Studies}
\label{appendix-ablation}
In this section, we provide extensive ablation studies on the context window size, the positional encoding, and the division compression strategy.

\noindent \textbf{Influence of the Context Window Size.} 
Context modeling is critical for compression because it contributes to both the transform coding and entropy model. We conduct ablation studies on the context capacity where the context window length $L$ is set to $2048, 512, 128$ and $16$. The original LocoMoco model with $L=1024$ is used as a baseline. As shown in Table \ref{tab:appendix-ablation}, reducing the context window size $L$ results in performance degradation. For example, reducing $L$ from 1024 to 512 leads to a performance drop of around 4\%.
These results confirm the significance of our main proposal of utilizing the long-context modeling to model the strong dependencies for 3DGS compression. While enlarging the context window for $L=2048$ provides further expanded receptive fields and yields a 1.79\% better rate-distortion performance, this improvement is marginal relative to the computational resource consumption. Consequently, we choose $L=1024$ as our default configuration for a more favorable balance between compression performance and efficiency.

\noindent \textbf{Effectiveness of the DGCNN Positional Encoding.} Positional encoding contributes significantly to the attention modules in the transform coding network and entropy model. The LocoMoco adopts a DGCNN for positional encoding.
To evaluate the effectiveness of this positional encoding strategy, we replace all DGCNN layers in our attention blocks with the plain sinusoidal positional encoding  \cite{vaswani2017attention}. As shown in Table \ref{tab:appendix-ablation}, this variant, named Sinusoidal PE, leads to a 7.93\% degradation in the rate-distortion performance. This verifies the effectiveness of DGCNN. Compared to plain sinusoidal positional encoding, DGCNN dynamically captures local and global geometric information, which contributes to an effective attention module.

In addition to the 3D coordinates $\hat{\boldsymbol{\mu}}$, we explore the potential benefit of incorporating the primitives' features (i.e., the input of the attention module) into the input of the DGCNN. Given that the DGCNN module imposes no restriction on its input dimension, a straightforward option is to concatenate the features alongside the coordinates $\hat{\boldsymbol{\mu}}$ as a joint input to the DGCNN. However, as shown in Table \ref{tab:appendix-ablation}, this modification does not yield a significant improvement. Note that the inherent design of the attention mechanism has already facilitated dense feature interactions across the representation space. Therefore, introducing these features into the positional encoding module just duplicates this interaction, and provides marginal gains to the capability of the model.

\noindent \textbf{Impact of the division compression strategy.} 
In our LocoMoco, we adopt a division network that partitions the 3DGS color attributes into two groups: a variation-sensitive subset that is encoded by lossless compression to preserve high-fidelity reconstruction, and a complementary subset that is encoded using lossy compression for better storage efficiency. To verify the effectiveness of this hybrid compression strategy, we establish two alternative baselines that adopt either fully lossless or lossy color attribute compression. For these two baselines, we discard the division network and train a fully lossy/lossless color compression model, while keeping the network architecture and all other settings consistent with the original LocoMoco model. 

The rate-distortion performance evaluated on the DL3DV-GS dataset is provided in Table \ref{tab:appendix-hybrid-lossy}. As shown in the table, although the lossy color compression baseline reduces the storage overhead by encoding all color attributes to latent space, it bears a significant visual quality drop of more than 1 dB. Consequently, this variant is more suitable for applications where the compression ratio is more important. Meanwhile, the lossless color compression baseline yields a marginal visual quality improvement of 0.18 dB while its storage size is nearly doubled, which is also suboptimal in terms of rate-distortion performance.
These results demonstrate that the hybrid lossy-lossless compression strategy achieves a superior balance between the reconstruction fidelity and storage efficiency compared to the lossy or lossless compression methods. 

\begin{table}[t!]
    \centering
    \caption{Ablation study results on the division compression strategy, evaluated on DL3DV-GS. Comparison between the original LocoMoco model and baselines with fully lossy/lossless color compression are provided.}
    \resizebox{\linewidth}{!}{
    \begin{tabular}{ccc}
    \hline
    Method & PSNR (dB) & Size (MB) \\
    \hline
    LocoMoco (Ours) & 29.12 & 19.92 \\
    \hline
    Lossy Color Compression   & 27.38 & 13.20 \\
    Lossless Color Compression & 29.30 & 36.49 \\
    \hline
    \end{tabular}
    }%
    \label{tab:appendix-hybrid-lossy}
\end{table}

\end{document}